\documentclass[11pt]{article}
\usepackage{coling2020}
\usepackage{latexsym}
\usepackage{graphicx}
\usepackage{caption}
\usepackage{amsmath}
\usepackage{amssymb}
\usepackage{pgfplots}
\pgfplotsset{compat=newest}
\usepackage{comment}
\usepackage{times}
\usepackage{url}

\usepackage{rotating}
\usepackage{array,multirow}
\usepackage[hidelinks]{hyperref}
\usepackage{arydshln}
\usepackage{pbox}

\newcommand{\specialcell}[2][l]{%
  \begin{tabular}[#1]{@{}l@{}}#2\end{tabular}
}

\makeatletter
\newcommand{\setword}[2]{%
  \phantomsection
  #1\def\@currentlabel{\unexpanded{#1}}\label{#2}%
}
\makeatother

\newcommand{\STAB}[1]{\begin{tabular}{@{}c@{}}#1\end{tabular}}
\newcommand{\blue}[1]{\textcolor{blue}{#1}}
\newcommand{\red}[1]{\textcolor{red}{#1}}

\newcommand{\Table}[1]{Table~\ref{#1}}

\title{\textit{Pick a Fight} or \textit{Bite your Tongue}:\\ Investigation of Gender Differences in Idiomatic Language Usage}

\author{
	Ella Rabinovich$^{1}$\quad\quad\quad
	Hila Gonen$^{2}$ \quad\quad\quad
	Suzanne Stevenson$^{1}$ 
	\vspace{0.1cm} \\
	$^{1}$Dept. of Computer Science, University of Toronto, Canada \\
	$^{2}$Dept. of Computer Science, Bar-Ilan University, Israel \\
	\texttt{\{ella,hilagnn\}@gmail.com} \\
	\texttt{suzanne@cs.toronto.edu}
}
\date{}

\begin{document}
\maketitle
\begin{abstract}
A large body of research on gender-linked language has established foundations regarding cross-gender differences in lexical, emotional, and topical preferences, along with their sociological underpinnings. We compile a novel, large and diverse corpus of spontaneous linguistic productions annotated with speakers' gender, and perform a first large-scale empirical study of distinctions in the usage of \textit{figurative language} between male and female authors. Our analyses suggest that (1) idiomatic choices reflect gender-specific lexical and semantic preferences in general language, (2) men's and women's idiomatic usages express higher emotion than their literal language, with detectable, albeit more subtle, differences between male and female authors along the dimension of dominance compared to similar distinctions in their literal utterances, and (3) contextual analysis of idiomatic expressions reveals considerable differences, reflecting subtle divergences in usage environments, shaped by cross-gender communication styles and semantic biases.
\end{abstract}

%
%
\blfootnote{
%
%
%
%
 \hspace{-0.65cm}  
 This work is licensed under a Creative Commons Attribution 4.0 International Licence. Licence details: \url{http://creativecommons.org/licenses/by/4.0/}.
}

\section{Introduction}
\label{sec:intro}


Distinctions in language usage between men and women open a window into the differences in their perception of the world around them. 
Over the years, these differences have been studied extensively, both in empirical linguistics \cite{lakoff1973language,labov1990intersection,coates2011language}, and using data-driven computational approaches (e.g., \newcite{koppel2002automatically}, \newcite{schler2006effects}, \newcite{mihalcea2016men}).  Striking cross-gender deviations in lexical and syntactic properties have been identified, igniting debate as to their underlying psycholinguistic and sociological interpretations \cite{lakoff1973language,cameron1988lakoff,bergvall2014rethinking}.
In contrast to the study of literal linguistic utterances, only sparse prior research exists on gender differences in \textit{figurative language} (for review see \newcite{colston2004figurative}).  Figurative language is a common communicative device in which utterances use words in ways that deviate from their literal interpretation (e.g., \newcite{nunberg1994idioms}), including a range of devices, such as 
irony, sarcasm, personification, and idiomatic expressions.
Various communicative motivations for non-literal language have been proposed, such as 
distinguishing different intensities of the same emotion, and communicating ideas and subjective experiences  more vividly
(for a review, see \newcite{gibbs1994poetics}). Indeed, figurative language has been suggested to reflect the way that concepts are structured within human thought (e.g., \newcite{lakoff2008metaphors}). Consequently, differences in the way men and women perceive experiences and emotions give rise to a plausible hypothesis that male (M) and female (F) authors differ in their preferences in the choice of non-literal language; however, there has been no large-scale computational study conducted on the distinctions in such usage patterns.

This gap in the research stems in part from the sparsity of large-scale corpora of linguistic productions employing figurative language, annotated for the gender of a speaker. The increasing popularity of online discussion platforms offers an unprecedented opportunity to address this gap, by harvesting information from texts authored by hundreds of thousands of authors. In this study we introduce GenderReddit -- a large and diverse corpus of linguistic productions of men and women collected from the Reddit\footnote{\url{https://www.reddit.com/}.} platform; each author in GenderReddit is annotated with their (self-reported) binary gender.\footnote{Although gender can be viewed as a continuum rather than a binary variable, for the sake of simplicity, we limit this study to the two most prominent gender markers in Reddit: male and female.} With over $100$M posts and $3.4$B words by nearly $30$K distinct authors, this dataset introduces a unique opportunity to study the language of men and women at scale.  Here, we specifically investigate the use of figurative expressions in this new dataset, thereby bridging the gap in empirical research in this field.


We suggest and empirically quantify factors that drive gender-related preferences in non-literal language, focusing on one of the most pervasive types of figurative utterances: \textit{idiomatic expressions}. Idioms are phrases, such as `on cloud nine', `pick a fight', or `eat someone alive', 
whose meaning is not (entirely) deducible from that of the literal words that comprise them. Aiming at identifying similarities and differences in male and female idiomatic language, we focus on three research questions:
\begin{enumerate}
\vspace{-2mm}
\item [\setword{\textbf{Q1}}{question:q1}:] Do gender-specific lexical preferences affect our choices of idiomatic expressions, considering both the surface (explicit) form and the underlying meaning of an idiom?
\vspace{-2mm}
\item [\setword{\textbf{Q2}}{question:q2}:] To what extent do the choices of idioms differ between men and women along the three dimensions of emotion (valence, arousal, and dominance)?
\vspace{-2mm}
\item [\setword{\textbf{Q3}}{question:q3}:] What are the differences in contextual usage of idiomatic expressions between the two genders?
\end{enumerate}
\vspace{-2mm}
\noindent
In investigating these questions, we find distinctions in the preferences of male and female authors along all three of lexical, emotional, and contextual dimensions of idioms. 
The contribution of this work is, therefore, twofold: First, we collect and make available a large and diverse corpus of spontaneous linguistic productions by male and female authors, facilitating further research in this field. Second, we conduct the first large-scale computational study on gender-specific preferences in idiomatic language, corroborating assumptions regarding these differences in the psycholinguistic literature.\footnote{All data and code are available at \url{https://github.com/ellarabi/gender-idiomatic-language}.}

\section{Background and Related Work}
\label{sec:related-work}

A large body of work exists on identifying and interpreting differences in the language of men and women, focusing on both topical and stylistic distinctions between the two genders \cite{lakoff1973language,holmes1984women,labov1990intersection,holmes1990hedges}. The connection between gender and language has  also been studied in the field of sociolinguistics \cite{eckert2013language,lindsey2015gender,coates2015women}, suggesting (not uncontroversially) that men tend to use linguistic forms and a communicative style reflecting higher tendency to dominance, engagement, and control, while women are more likely to employ positive emotions. \newcite{thelwall2019she} studied cross-gender participation in various topical threads on Reddit, observing significantly higher engagement of male authors in gaming, technology, and sports, while higher female participation was found in subreddits on personal and domestic advice. Gender-linked distinctions in topical and lexical preferences were also highlighted by \newcite{schwartz2013personality}, who demonstrated significant correlation of the presence of certain LIWC categories in a text and its author's gender), as well as by \newcite{rangel2017overview}, who reported up to $80$\% accuracy on a binary gender classification task using ngrams of words.

Recently, the computational investigation of male and female language has been prolific, yielding an empirical foundation for the theoretically-motivated hypotheses on characteristics of the two linguistic varieties. Automatic classification of an author's gender has shed much light on the manifestation of gender-specific traits in language, including differences in the usage of pronouns, numerals, emotion markers, and intensifiers (e.g., `definitely', `absolutely') \cite{koppel2002automatically,argamon2003gender,schler2006effects,rabinovich2017personalized}. \newcite{mihalcea2016men} addressed the task of gender identification as a word-sense disambiguation scenario, and demonstrated that an author's gender can be accurately detected from the contextual environments of a few hundred common words in the language. Marked gender differences on emotional dimensions were shown by \newcite{mohammad2011tracking}, who found that women tend to use words from the joy--sadness axis, whereas men prefer terms from the fear--trust axis. \newcite{memon2019detecting} found gender differences in perception of emotion, reporting that women rate the degree of emotion in speech fragments higher than men.

Studies on gender-related differences in figurative language are relatively sparse (occasionally conducted in a laboratory experimental setting), with no large-scale computational investigation on the differences between M and F authors. Research has shown that men use more sarcasm and irony than women in spoken communication \cite{gibbs2000irony}, even when the risk of misinterpretation is equally estimated by men and women \cite{colston2004gender}. \newcite{link2005men} found  gender differences in preferences in non-literal language emotional communication, reporting that men used more figurative language in description of negative  (vs.\ positive) emotions, while no significant difference was found in female language. Our work is the first large-scale computational study on gender differences in figurative language, focusing on idioms as a pervasive communicative device.
\section{Dataset Collection and Preprocessing}
\label{sec:datasets}

\subsection{Linguistic Productions Annotated with Author Gender}
We collected a large dataset of  
linguistic utterances in English by male and female authors from the Reddit discussion platforms. As of May 2020, Reddit was ranked as the $19$th most visited website in the world, with over $430$M active users, $1.2$M topical threads (subreddits), and over $70$\% of its user base coming from English-speaking countries. 
Many subreddits allow (and encourage) their subscribers to specify a meta-property (called a `flair', a textual tag) that customizes their presence within the subreddit. We identified a set of subreddits, such as `r/askmen' and `r/askwomen', where authors commonly self-report their gender, and extracted a set of unique user-ids of authors who did so. Using these extracted ids,
we collected the entire digital footprint of $13,630$ male and $16,182$ female users from the Reddit discussion platforms, spanning years 2005--2020, resulting in over $100$M posts, comprising spontaneous utterances by $29,812$ unique authors in over $100$K topical threads.\footnote{Because self-described men are represented more than self-described women on reddit, we randomly downsampled all posts by M authors, balancing the amount of tokens with those by F authors, to ease comparative analyses in this work. The full dataset, including additional posts by M authors, is available upon request.} The ample size of the corpus, as well as its diverse nature, facilitate the analysis of the similarities and differences in the usage of figurative language by gender, a topic that has been understudied computationally thus far.

\subsection{Idiomatic Expressions}
Drawing on the Cambridge Dictionary of Idiomatic Expressions\footnote{\url{https://dictionary.cambridge.org/dictionary/english/}} and additional online resources, we collected a preliminary set of idiomatic expressions, including both the surface forms and their definitions (literal meanings);
e.g., `at odds': `in conflict or at variance, not agreeing with each other'; `sit on the fence': `to avoid taking sides in a discussion or argument'. This process yielded a large and diverse set of idioms, covering various semantic domains (emotions, relationships, nature, etc). 

Idioms may differ in actual use from their canonical form -- the form that appears in a dictionary. In some cases, virtually no variation is allowed without loss of the idiomatic sense, while other cases allow extensive types of variation. For the purpose of this study we focused on generalizing over two common types of variation~\cite{spasic2017idiom}: (1) verb inflection (e.g., `pick a fight' may surface as `picking a fight' or `picked a fight') and (2) open slots in expressions involving indefinite pronouns (e.g., `swallow \textit{one's} pride', `mean the world to \textit{someone}'), which generally are used with the indefinite pronoun substituted by a personal pronoun (e.g., `swallow \textit{her} pride', `mean the world to \textit{me}'). We refrain from handling additional variations -- e.g., filling in open slots with noun phrases
-- due to the difficulty of accurate extraction and the lower frequency of these cases. We detail our approach to handling these two types of variation in Section~\ref{sec:expanding}.

Certain idiomatic expressions are inevitably ambiguous since they may be interpreted both literally and idiomatically; e.g., `let the cat out of the bag' idiomatically means to reveal a secret, but can literally refer to the stated action. However, while some of our idioms have a \textit{possible} literal reading, typically many fewer have a \textit{common} literal reading.
(Our findings are consistent with those of earlier work; for example, \cite{fazly2009unsupervised} found that for 2/3 of the potentially-idiomatic expressions in their token dataset---i.e., phrases that could be used with either an idiomatic or literal meaning---over 75\% of their usages were in an idiomatic reading.) For example, we find frequent literal usages of `ball of fire' (typically referring to a star), while `raise eyebrows', although having a readily available literal meaning, is much less likely to be used in a non-idiomatic manner. Focusing on the study of idiomatic usages, we detail our approach to filtering out expressions with a common literal meaning in Section~\ref{sec:identify-literal}.

Following the extraction and filtering steps detailed below, the final set comprises $527$ unique idioms (i.e., representing all variants of an expression under its single canonical form, covering a total of $835$ surface forms), along with their definitions. See \Table{tbl:dataset}, and list of idioms in supplemental materials.

\begin{table}[h]
\centering
\begin{tabular}{lrrrrrrr}
gender & authors & subreddits & posts & tokens & \# of unique idioms (variants) & \# of idiom instances \\ \hline
M & 13,630 & 88,668 & 57.3M & 1.7B & 527 (835) & 335,572 \\
F & 16,182 & 70,943 & 42.7M & 1.7B & 527 (835) & 328,830 \\
\end{tabular}
\vspace{-0.2cm}
\caption{Dataset statistics.}
\label{tbl:dataset}
\end{table}

\subsubsection{Expanding Idiomatic Expressions with Additional Surface Forms}
\label{sec:expanding}

For idioms with verb constituents, we exhaustively expanded the canonical idiom form with all possible verb tenses.\footnote{We used the \href{https://github.com/bjascob/pyinflect}{\texttt{pyinflect}} module to retrieve all verb tense forms.}  
For idioms with indefinite pronouns as open slots, we automatically substituted the indefinite pronouns with the exhaustive set of personal pronouns (e.g., `she', `me', `you', `my', `his', etc.).  
While achieving full coverage, this fully-automated approach is inferior in precision, generating ungrammatical forms (e.g., `begin to saw the light' ) or extremely unlikely ones (`holding their horses'). Aiming at collecting a viable set of idioms, we restricted the set of automatically generated individual surface forms to those exceeding $50$ occurrences in our data.
%
In all our further analyses (in Section~\ref{sec:model}), we consider all the extracted surface forms of an expression as an instance of the unique idiom, which we refer to by its canonical form; as an example, the cumulative counts of `pick a fight' include instances of `pick a fight', `picked a fight', `picks a fight', etc.

\subsubsection{Filtering Out Expressions with a Common Literal Reading}
\label{sec:identify-literal}

Because our goal is to analyze idiomatic language use, we aim to automatically remove from our list of idioms those with a high degree of \textit{literality} -- i.e., a high potential for literal interpretation.\footnote{Removing the \textit{instances} in the corpus of idiomatic expressions that are used literally would be a noisy alternative, since token-based identification of idioms is an open research problem. For example, \newcite{fazly2009unsupervised} propose a method for identifying literal usages of idioms, but it does not generalize to the range of idiomatic forms in our list, since it depends on syntactic properties of idioms of a particular form.  We take the more conservative approach of simply removing from consideration the idiomatic expressions that are likely to be commonly used in their literal interpretation.}  Relevant work has focused on the related property of compositionality -- the extent to which an idiomatic interpretation is formed from the component words (e.g., \newcite{mccarthy2007detecting}, \newcite{salehi2015word}).  While this property is related to literality (Spearman's $\rho{=}0.62$ in human ratings, \newcite{nordmann2014familiarity}) it is not identical, since highly decomposable idioms may not actually occur frequently in their literal meaning.  However, because the method of \newcite{salehi2015word} uses word embeddings to detect compositionality (comparing the embedding for the entire idiom to the embeddings for its component words), it is a suitable approximation to literality: i.e., the frequent use of the idiomatic form in its literal meaning entails that the semantic representation of the idiom will be closer to those of the individual words. 
Specifically, we adopt as our measure of literality the average similarity between the embedding of an idiom as a single token and the embeddings of its individual constituents. 

\Table{tbl:literacy} presents some idioms with highest and lowest literality scores.
Interestingly, `water under the bridge' was assigned a very low literality score, indicating that despite its readily-available literal interpretation, this expression is most predominantly used in its idiomatic meaning in our dataset. By contrast, the relatively high score assigned to `apples and oranges' implies a significant ratio of literal contexts of this expression. We, therefore, imposed an (empirically determined) strict threshold of $0.25$, filtering out idiomatic expressions exceeding this literality score from our analysis.

\begin{table}[h]
\centering
\begin{tabular}{p{4cm}r|p{4cm}r}
idiom & score & idiom & score \\ \hline
wooden spoon & 0.600        & firing on all cylinders & 0.015 \\
on the bandwagon & 0.593    & through thick and thin & 0.042 \\
black and blue & 0.498      & water under the bridge & 0.046 \\
apples and oranges & 0.441  & reading between the lines & 0.063 \\
ball of fire & 0.414        & means the world to me & 0.111 \\
\end{tabular}
\caption{Automatically inferred degree of literality for example idioms: high (left) and low (right).}
\label{tbl:literacy}
\end{table}

\section{Differences Across Genders in Figurative Language Usage Patterns}
\label{sec:model}


In each subsection below, we define and apply quantitative and qualitative analyses that address each of the three research questions, \textbf{Q1}--\textbf{Q3}, 
finding considerable differences between the two genders.

Before turning to our specific research questions, we first verify that the overall usage patterns of idioms differ between our male and female subcorpora.  Briefly, we calculate two probability distributions for M and F usage over the list of $527$ idiomatic expressions, and compare them using Jensen-Shannon divergence (JSD), obtaining a value of $0.140$.  We find that this value differs significantly (p\textless$1.0e{-}6$) from the average JSD of 500 random splits \textit{within} M usages (mean JSD=$0.038$) or F usages (mean JSD=$0.035$). Thus there are indeed significant cross-gender distinctions in the usage of idiomatic expressions, compared to variation in random samples within the same gender.

\subsection{Lexical Preferences in the Usage of Idiomatic Expressions by Men and Women}

We hypothesize that the choices of speakers in the domain of figurative language mirror their general linguistic preferences, in two ways.  First, we suggest that gender-related lexical preferences will `shine-through' men's and women's idiomatic choices; 
that is, words that are generally characteristic of female (male) language will be more common in the idiomatic forms women (men) choose, despite the non-literal interpretation of those words.
Second, we also expect that the underlying meaning of idioms used more by F will reflect semantic biases associated with the language of women, and similarly for men.

We address these hypotheses by extracting lexical markers of M and F language in our corpus, using the \textit{log-odds ratio} with informative Dirichlet prior \cite{monroe2008fightin}.  This statistical method assumes two subcorpora of two language varieties to be compared, along with a neutral background corpus comprising a balanced mix of the two. The approach then assigns each token (or a phrase treated as a single token) a score, reflecting the strength of its association with one of the language varieties or the other.  

\begin{figure}[hbt]
\centering
\includegraphics[width=16cm]{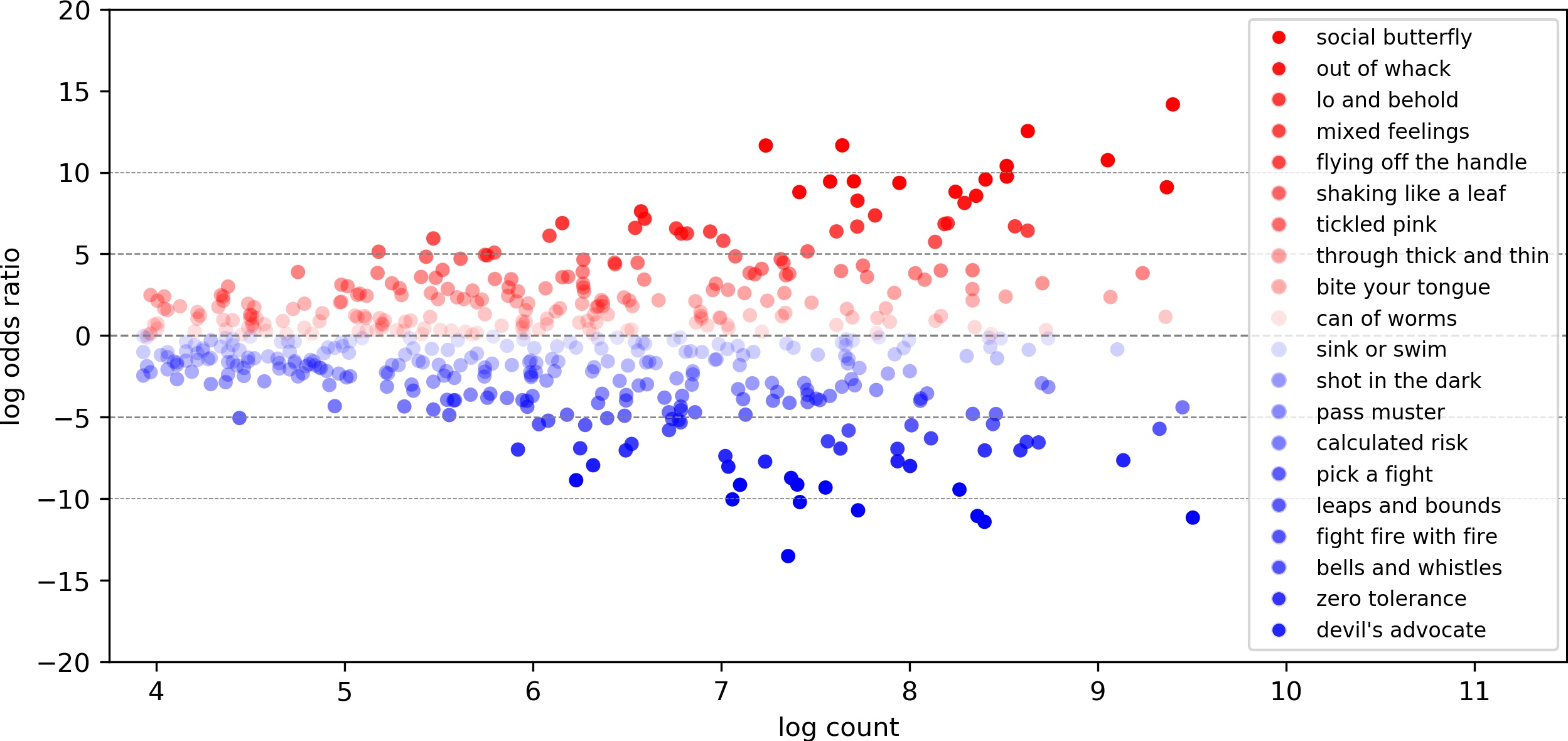}
\caption{$gScore$ of idioms as a function of log count in the corpus. Greater dispersion of points (on the right-hand side) is indicative of more significant differences in the usage of individual expressions between the two genders. We leave in-depth analysis of these divergences for future work.}
\label{fig:gender-scatter}
\end{figure}

We first use this method to assign a gender-association score, $gScore(i)$---negative for M and positive for F---to each idiom $i$ of our $527$ idioms, as the log-odds ratio of $i$ in our M and F subcorpora.  Figure~\ref{fig:gender-scatter} presents the distribution of $gScore(i)$ as a function of $i$'s log-scaled counts in M and F language; the legend on the right presents a sample of idioms depicted in the graph. Positive scores (in red) refer to expressions used more by women and negative scores (in blue) to those used more by men; the absolute value of a score is the strength of an idiom's association with one of the genders.  Note that higher counts in the data yield a higher dispersion of markers, suggesting that the significance of this metric benefits from a very large number of usage examples such as can be provided by a corpus such as ours. 

This figure reveals interesting observations about gender and figurative language.  First, the only-slightly unbalanced distribution ($284$ M vs.\ $243$ F idioms) is nearly symmetric along the y-axis, implying similar patterns of the spread of gendered association of idioms in the two subcorpora. Intriguingly, many of the examples to the right in Figure~\ref{fig:gender-scatter} suggest an intuitive correlation between M and F preferences of semantic domains, and the lexical terms making up the form of their idiomatic choices (e.g., `social butterfly' and `mixed feelings' for women, and `pick a fight' and `fight fire with fire' for men).

Table~\ref{tbl:gender-idioms} reports additional examples of idioms most associated with M and F language, along with their counts (recall our M and F corpora are equal-sized samples), further supporting the observation that cross-gender semantic propensities shape our choices in the form of figurative language: idioms using terms from the domain of sports and warfare are more pervasive in M language, while expressions using words related to housekeeping and family are more frequent in F choices. These findings are in line with insights from analysis of cross-gender topical preferences on Reddit \cite{thelwall2019she}. 

\begin{table}[h]
\centering
\begin{tabular}{p{4cm}rr|p{4cm}rr}
idiom & count(M) & count(F) & idiom & count(M) & count(F) \\ \hline
across the board & 7881 & 5634 & over the moon & 781 & 2258 \\
level playing field & 1151 & 501 & stick to your guns & 803 & 1480 \\
under fire & 1105 & 495 & head over heels in love & 201 & 537 \\
front runner & 781 & 344 & on pins and needles & 287 & 615 \\
make a killing & 1317 & 600 & a shoulder to cry on & 393 & 729 \\ 
firing on all cylinders & 399 & 121 & throw in the towel & 998 & 1360 \\
reinvent the wheel & 949 & 445 & recipe for disaster & 1575 & 1977 \\
\end{tabular}
\caption{Examples for idiom choices used more by men (left) and by women (right).}
\label{tbl:gender-idioms}
\end{table}

On the other hand, there are exceptions to this pattern, such as `stick to your guns', which is associated more with female speakers.  We observe that the explicit language of the \textit{surface form} of an idiom is not the only factor driving gender-based choices; a plausible assumption would be that the idiomatic interpretation affects these preferences as well.  For example, the expression `on the same wavelength' is more common in women's language (counts of $345$ for M and $690$ for F) due to the fact that this expression's \textit{underlying meaning} is better associated with topics more commonly discussed by women, even if the component words of the idiom are not.

We have thus seen qualitative evidence that M/F preferences in the use of an idiom may be influenced by both the lexical makeup of the surface form, and the domain of its underlying meaning.
We next quantify this intuition by measuring the correlation of the gender score assigned to an idiom with that of the words in its surface form and in its definition. Formally, let $gScore(w)$ be the gender score assigned to each word $w$ in the lexicon by the log-odds ratio analysis on the M/F subcorpora, analogous to $gScore(i)$ (as above) for idioms $i$. We further define $gScore^S(i)$ and $gScore^D(i)$ to be the scores assigned to the surface form and the definition of idiom $i$, by averaging over the $gScore$s of individual words $w$ in the set of words in the idiom surface-form $W^S(i)$ and its definition $W^D(i)$, respectively:
\begin{equation}
gScore^S(i) = \dfrac {\sum gScore(w_k)} {|W^S(i)|}, \quad\, w_k \in W^S(i)
\label{eq:gender-scores-1}
\end{equation}
\vspace{-0.1cm}
\begin{equation}
gScore^D(i) = \dfrac {\sum gScore(w_k)} {|W^D(i)|}, \quad w_k \in W^D(i)
\label{eq:gender-scores-1}
\end{equation}

Now to measure how much M/F choice of an idiom is reflective of their preferences for words in its surface form, we take the correlation of $gScore(i)$ and $gScore^S(i)$ over our $527$ idioms.  Similarly, we assess how much idiom choice is reflective of the semantic domain of the underlying meaning, by the correlation of $gScore(i)$ and $gScore^D(i)$. As expected, Spearman's $\rho$ resulted in a modest positive correlation between $gScore(i)$ and $gScore^S(i)$  (r{=}0.304, p\text{-}val{=}2.3e\text{-}13), as well as between $gScore(i)$ and $gScore^D(i)$ (r{=}0.314, p\text{-}val{=}9.2e\text{-}15).  These results support our suggestion that gender preferences in non-literal language are driven (to some extent) by M and F natural tendencies in both their lexical choices and semantic preferences, providing a positive answer to our \ref{question:q1}.

\subsection{Emotional Distinctions in Idiom Preferences of Men and Women}

A common way to study emotions in the psycholinguistic literature 
groups affective states into a few major dimensions. The Valence-Arousal-Dominance (VAD) representation has been widely used to conceptualize an individual's emotional spectrum, where \textit{valence} refers to the degree of positiveness of the affect, \textit{arousal} to the degree of its intensity, and \textit{dominance} represents the degree of feeling in control \cite{bradley1994measuring}. For example, the word `fabulous' is ranked high on the valence dimension, while `deceptive' is rated low. Examples for words with high and low arousal include `explosive' (high) and `natural' (low); contrasting dominance ratings are found for `masterful' (high) and `weak' (low). Previous studies \cite{burriss2007psychophysiological,hoffman2008empathy,thelwall2010data} have suggested that women are more likely than men to employ positive emotions, while men exhibit higher tendency to dominance, engagement, and control (although see \newcite{park2016women} for an alternative finding). In this study we test if gender-related preferences in figurative language reveal differences along these emotional dimensions. Specifically, we test if the distributions of the three emotion variables across idiom usages differ between men and women, where the values of the emotional dimensions are based on the underlying meanings (definitions) of the idiomatic expressions.

\subsubsection{Assessing the Emotional Dimensions of Idiom Definitions}

A large dataset of human rankings of VAD for $20,000$ English words was recently released by \newcite{mohammad2018obtaining}, where each word is assigned a value for each of the three dimensions on a $0\text{-}1$ scale.
Because we aim at estimating the affective variables of phrases (idiom definitions), rather than individual words, we must automatically infer the affective ratings of phrases using those of individual words, as follows.

Word embedding spaces have been shown to capture variability in emotional dimensions closely corresponding to valence, arousal, and dominance \cite{hollis2016principals}, implying that such semantic representations carry over information useful for the task of emotional affect assessment. Therefore, we exploit affective dimension ratings assigned to individual words for supervision in extracting ratings of sentences. We use the model introduced by \newcite{reimers2019sentence} for producing word- and sentence-embeddings using Siamese BERT-Networks, 
thereby obtaining semantic representations for the $20,000$ words in \newcite{mohammad2018obtaining} as well as for the $527$ idiom definitions in this study.

Beta regression models\footnote{An alternative to linear regression in case where the dependent variable is a proportion (0\text{-}1 range).} were then trained to predict VAD scores from the embeddings for the $20$K individual words, obtaining Pearson's correlations of $0.85$ (V), $0.78$ (A), and $0.81$ (D) with the human annotated ratings on a held-out set of $1000$ words. The three trained models were then used to infer VAD measurements for the entire set of definitions (sentence embeddings). Table~\ref{tbl:vad-definitions} presents a sample of idioms that were assigned contrasting VAD values. Note that the three dimensions do not necessarily correlate with each other, e.g., `under fire' has a low dominance but a high arousal value.

\begin{table}[h]
\centering
\resizebox{\textwidth}{!}{  
\begin{tabular}{cllr}
& idiom & definition & score \\ \hline
\multirow{4}{*}{\STAB{\rotatebox[origin=c]{0}{V}}} 
& on cloud nine & to be extremely happy & 0.987 \\
& make a killing & to have had great financial success & 0.800 \\ \cdashline{2-4}
& reduced to tears & when your behaviour or attitude makes someone cry & 0.039 \\
& under fire & when someone is being attacked and criticized heavily & 0.002 \\ \hline
\multirow{4}{*}{\STAB{\rotatebox[origin=c]{0}{A}}} 
& jump out of one's skin & to be extremely surprised, frightened or shocked & 0.950 \\
& under fire & when someone is being attacked and criticized heavily & 0.941 \\ \cdashline{2-4}
& scratch the surface & to deal with only a small part of a problem & 0.360 \\
& at the eleventh hour & the last moment or almost too late & 0.356 \\ \hline
\multirow{4}{*}{\STAB{\rotatebox[origin=c]{0}{D}}}
& put someone's heart into & to  be very enthusiastic and invest much energy in something & 0.932  \\
& fight tooth and nail & to fight fiercely, with energy and determination & 0.892 \\ \cdashline{2-4}
& tongue-tied & difficulty in expressing yourself due to embarrassment & 0.267 \\
& blue in the face & to be exhausted and speechless & 0.251 \\
\end{tabular}
}
\vspace{-0.1cm}
\caption{Automatic induction of valence (V), arousal (A) and dominance (D) for idiom definitions. 
}
\label{tbl:vad-definitions}
\end{table}

\subsubsection{Quantifying VAD Differences in Male and Female Usage of Idioms}

Next we estimate the differences across the three emotion dimensions in the idiomatic preferences of male and female authors. Specifically, all idiomatic expression instances in the two sets of posts (M and F) are assigned a three-dimensional value---V, A, and D---based on their underlying definition. For each gender, this results in three vectors (one per emotional dimension) of the list of all idiom usages in that gender's posts.  
Thus, for example, the V vector for F has a list of values for valence for each idiom, with the V value for idiom $i$ occurring $count(i)$ times in the vector, where $count(i)$ is the number of occurrences of idiom $i$ in the F corpus.
Wilcoxon ranksum statistical test is then applied to the M/F pairs of series of values for each of V, A, and D, testing for significant difference, and Cohen's $d$ calculated to indicate the magnitude of the effect.

Table~\ref{tbl:vad-differences} reports the results.  While only a slight (albeit significant) difference holds between M/F valence values, and virtually no distinction exists on the dimension of arousal, the dimension of dominance shows a greater difference, with men tending to use idiomatic expressions carrying a higher degree of feeling of control more frequently than women.  Figure~\ref{fig:dominance-density} illustrates the kernel density estimation of idiom dominance values: the small difference in idiom dominance is reflected in the slight right shift of the density function in men's idiom usages, compared to women’s.  Table~\ref{tbl:vad-examples} further presents a few example utterances including high- and low-dominance idioms in our corpus.  For comparison, we similarly computed the difference in dominance when measured on two equal-sized samples of spontaneous utterances (not including idiomatic expressions) by M and F authors, which were both found to have lower dominance values than for their idiomatic language. Moreover, the M/F non-idiomatic utterances have a larger Cohen's $d$ effect size (of $0.15$, compared to $0.08$ between the M/F samples of idioms).\footnote{While less distinct than the pattern with dominance, arousal was also somewhat lower in literal language for both genders and with a greater M/F effect size, while valence varied little in the two types of language.}  These collective analyses suggest that the emotional dimension of dominance, and to a lesser extent arousal, are heightened in idioms, confirming the role of figurative language in expressing emotional content \cite{gibbs1994poetics}.  Moreover, 
the generally more-dominant communicative style of men relative to women discussed in the literature \cite{eckert2013language,lindsey2015gender,coates2015women} remains in evidence in their figurative language, thereby shedding some light on our \ref{question:q2}.

\begin{table}
\begin{minipage}{0.45\linewidth}
\centering
\begin{tabular}{p{1.5cm}ccr}
dimension & \multicolumn{1}{c}{mean(M)} & \multicolumn{1}{c}{mean(F)} & \multicolumn{1}{c}{Cohen's $d$} \\ \hline
V*  & \textbf{0.444} & 0.440 & 0.02 \\
A   & 0.619 & 0.620 & 0.00 \\
D**  & \textbf{0.549} & 0.538 & 0.08 \\
\end{tabular}
\caption{Distinctions along the three VAD emotion dimensions in male vs.\ female idiom usages. Significant differences are marked by `*' (p\textless.01) and `**' (p\textless.001), and effect sizes are given by Cohen's $d$.}
\label{tbl:vad-differences}
\end{minipage} \hfill
\begin{minipage}{0.5\linewidth}
\centering
\includegraphics[width=7cm]{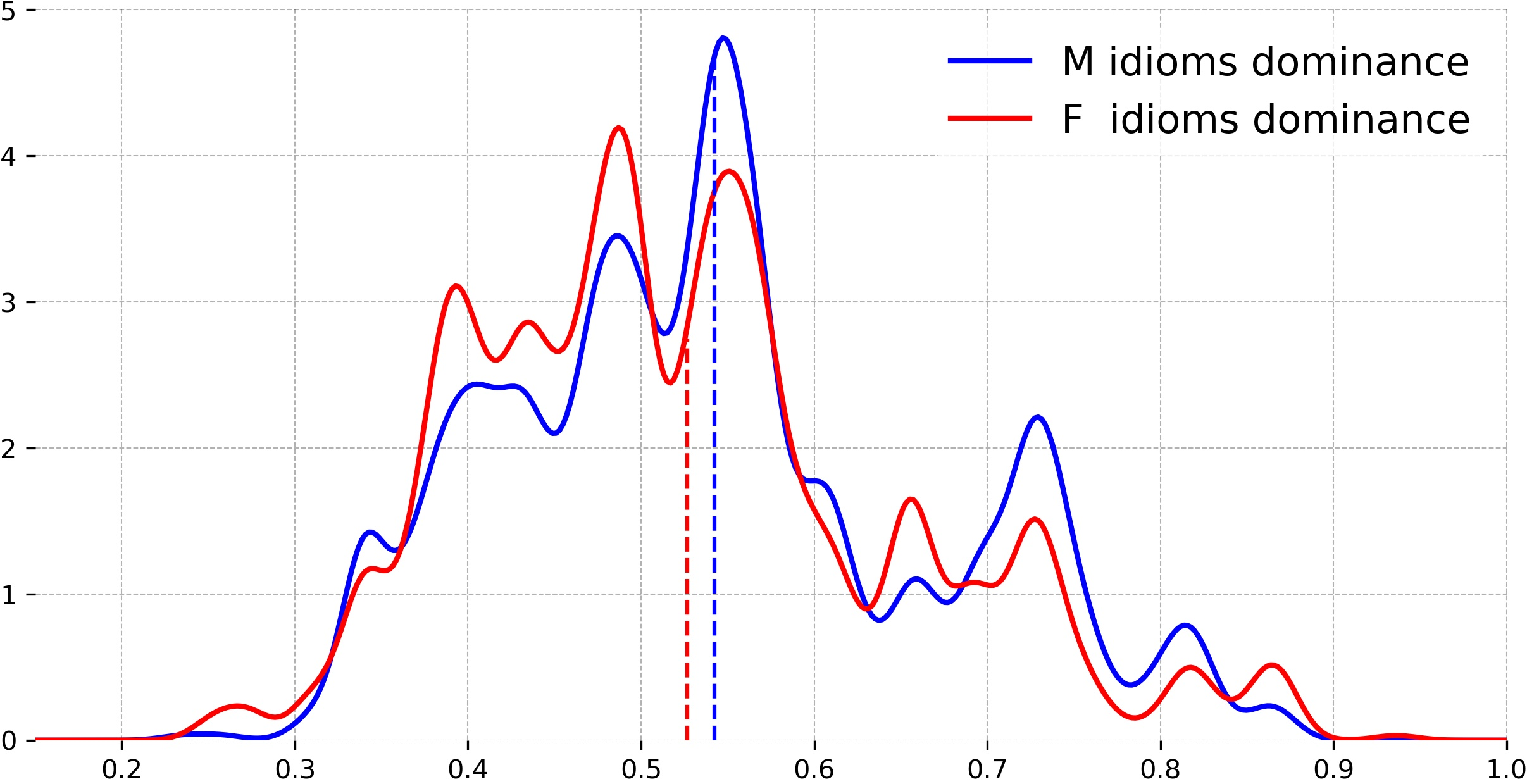}
\captionof{figure}{KDE of M vs.\ F idiom dominance. 
}
\label{fig:dominance-density}
\end{minipage}
\end{table}

\begin{table}[h]
\centering
\resizebox{\textwidth}{!}{  
\begin{tabular}{ccl}
D & gen. & \multicolumn{1}{c}{example usage} \\ \hline
\multirow{3}{*}{\STAB{\rotatebox[origin=c]{90}{high}}} 
& M & \specialcell{My brother and I gave our mother hell, yet she’d fight \textbf{tooth and nail} for us.} \\ \cdashline{2-3}
& M & \specialcell{We \textbf{came a long way} from Global Warming is a hoax to give up to excuses such as \\ ``we can't stop it.'' We sent a f**g man to the moon.} \\ \hline
\multirow{3}{*}{\STAB{\rotatebox[origin=c]{90}{low}}} 
& F & \specialcell{Let him learn how to take care of the baby on his own. You have a lot more experience, \\ so you'll have to \textbf{bite your tongue} sometimes. If it's not likely to injure the baby, let it go.} \\ \cdashline{2-3}
& F & \specialcell{You can talk till you're \textbf{blue in the face} and the person across the table will not have received \\ the message. Meanwhile, find some emotional support through friends and family.} \\
\end{tabular}
}
\vspace{-0.1cm}
\caption{Example usages of high- and low-dominance idioms taken verbatim from our corpus.}
\label{tbl:vad-examples}
\end{table}

\subsection{Qualitative Analysis of Differences in Contextual Usage of Idioms Across Genders}

Inspired by the large body of research showing differences in the way men and women use words in context, we aim to identify contextual deviations in the figurative language of the two genders, addressing \ref{question:q3}. Specifically, we ask if men and women employ idiomatic expressions similarly, or there are differences in the way they use them, conveying distinct shades of meanings of the same expression, that exhibit varying association strength with the gender of the author. Contemporary distributional semantics methodologies suggest a direct way to answer these questions. In order to reliably identify the similarities and differences in contextual usages of figurative language by men and women, we train two sets of embeddings---for M and F utterances---where each idiomatic expression is treated as a single token. Comparison of semantic neighborhoods within each embedding space provides a window into the subtle distinctions in usage patterns typical to the language of the two genders.

We trained two sets of embeddings (using the \href{https://radimrehurek.com/gensim/}{\texttt{gensim}} package), where all surface variants of an idiomatic expression (cf.~Section~\ref{sec:expanding}) were treated as a single token in the training data. Because our goal is to identify expressions with the most distinct contextual usages across the genders, we determine for each idiom
the similarity of their immediate neighbourhoods across the M and F sets of trained representations.  Using a variation of the approach of \newcite{gonen2020ctx}, we compute the gender-language similarity of an idiom by finding the intersection of its neighbors in semantic space across the two genders, weighting close neighbors higher than more distant ones, 
in a ranked list of the $100$ closest neighbours. Formally, we denote by $N^k_i(M)$ the ranked list of the $k$ most similar words to idiom $i$ in semantic space trained on male productions, and by $N^k_i(F)$ its counterpart list for embeddings trained on female language. A variation of the \textit{ranked-biased overlap} (RBO) metric \cite{webber2010similarity} is then applied to compute the similarity of the two lists. Assuming decreasing importance of list elements, this information-retrieval-inspired measurement assigns higher weight to words at the top of the list (nearest neighbors), and lower weights to more distant neighbors (at the bottom of the list):

\begin{equation}
simRBO(i) = \frac{1}{100} \sum_{k} \dfrac {|N^k_i(M) \cap N^k_i(F)|} {k}, \quad k \in \{1, .., 100\}
\label{eq:context-diffs}
\end{equation}



Table~\ref{tbl:neighbors} presents examples of the 
closest $10$ neighbors of
idioms 
with the lowest gender-language similarity
(as defined in Equation~\ref{eq:context-diffs}) across the two embedding spaces for M and F. We suggest that the observed patterns of differing neighbor sets reflect (to some extent) differences in communication style between the two genders -- e.g., women engage in more collaborative patterns, whereas men display a more competitive mode of interaction \cite{aries1996men}. Further examination of the highlighted differences raises a question regarding the extent to which this deviation can be explained by preferences in lexical choices in M vs.\ F language. As an example, the 
excessive frequency of `closure' in women's posts ($11$K in F vs.\ $7$K in M), in contrast to the opposite pattern for `negotiate' ($9$K in F vs.\ $13$K in M),
could explain why the former is a close neighbor of `bury the hatchet' in F embeddings, and the latter in M embeddings. However, this phenomenon does not hold for the M neighbour `settle' that is used more by women ($36$K in F vs.\ $32$K in M). We hypothesize that these general gender-related lexical preferences only partially explain the deviations in contextual usages observed in our analysis.
While our findings provide some answers in response to~\ref{question:q3}, we leave in-depth investigation of this question to future work.

\begin{table}[h]
\centering
\resizebox{\textwidth}{!}{  
\begin{tabular}{cc|cc|cc|cc}
\multicolumn{2}{c|}{\textbf{bury the hatchet}} & \multicolumn{2}{c|}{\textbf{let bygones be bygones}} & \multicolumn{2}{c|}{\textbf{tit for tat}} & \multicolumn{2}{c}{\textbf{nerves of steel}} \\ \hline
M & F & M & F & M & F & M & F \\ \hline
\blue{win}          & clarify           & \blue{surrender}      & \red{chillout}        & \blue{voluntary}      & petty                 & dedication        & patience \\
\blue{retire}       & compromise        & compromise            & \red{clarify}         & \blue{deliberate}     & \red{pointless}       & patience          & \red{willpower} \\
\blue{defeat}       & \red{chime}       & \blue{gtfo}           & \red{reconsider}      & \blue{disagreement}   & \red{childish}        & talent            & guts \\
\blue{negotiate}    & \red{closure}     & \blue{escalate}       & compromise            & constructive          & \red{trivial}         & skills            & dedication \\
\blue{resolve}      & \red{empathize}   & \blue{accept}         & \red{embrace}         & \blue{semantics}      & constructive          & guts              & talent \\
compromise          & \red{friendship}  & \blue{negotiate}      & \red{discuss}         & \blue{aggression}     & \red{logical}         & \blue{medal}      & skills \\
\blue{settle}       & \red{proceed}     & \blue{abandon}        & \red{misread}         & \blue{verbal}         & \red{mutually}        & \blue{charisma}   & \red{expertise} \\
\blue{confront}     & \red{sympathize}  & \blue{proceed}        & \red{ignorant}        & \blue{tactic}         & \red{fair}            & \blue{balls}      & \red{strength} \\
clarify             & \red{articulate}  & \blue{confront}       & \red{generalize}      & petty                 & \red{partnership}     & \blue{courage}    & \red{confidence} \\
\blue{beat}         & \red{argue}       & \blue{settle}         & \red{examine}         & \blue{strategy}       & \red{dishonest}       & \blue{feats}      & \red{hubs} \\

\end{tabular}
}
\vspace{-0.1cm}
\caption{Contextual usage differences in top-$10$ nearest neighbors of idiomatic expressions. 
}
\label{tbl:neighbors}
\end{table}

\section{Conclusions}
\label{sec:conclusions}
We compile a novel dataset of linguistic productions annotated with speakers' gender, and use it to conduct the first large-scale empirical study of cross-gender differences in preferences of idiomatic expressions -- along lexical, emotional, and contextual dimensions. The released dataset is likely to facilitate future exploratory activities in this field. Our future plans include further investigation of the distinctions in male vs.\ female productions, focusing on additional types of figurative language (e.g., irony, sarcasm and personification), a different genre (e.g., literature), or modality (e.g., spoken language). Additionally, our work focuses only on English. Conducting similar research with languages other than English, and comparing the findings, would be another interesting direction.

\section*{Acknowledgements}
We are thankful to Sivan Rabinovich, who laid out the intuition for main ideas presented in this work. We would like to thank Blair Armstrong and Barend Beekhuizen for their useful comments. We are also grateful to our anonymous reviewers for their constructive feedback. The research was supported by NSERC grant RGPIN-2017-06506 to Suzanne Stevenson.

\bibliographystyle{coling}
\bibliography{main}

\end{document}